\documentclass[10pt,twocolumn,letterpaper]{article}

\usepackage{verbatim}
\usepackage{iccv}
\usepackage{times}
\usepackage{epsfig}
\usepackage{graphicx}
\usepackage{amsmath}
\usepackage{amssymb}
\usepackage{adjustbox}
\usepackage{colortbl}

\usepackage{pifont}
\usepackage[square, sort, comma, numbers]{natbib}
\makeatletter
\@namedef{ver@everyshi.sty}{}
\makeatother
\usepackage{algorithm2e}
\usepackage[breaklinks=true,bookmarks=true,colorlinks=true,linkcolor=blue]{hyperref}

\iccvfinalcopy 

\def\iccvPaperID{10039} 

\ificcvfinal\fi

\newcommand\tick{\checkmark}

\def\*#1{\mathbf{#1}}
\newcommand\Real{\mathbb{R}}
\newcommand\va{\*{v}}
\newcommand\au{\*{a}}
\newcommand\expr{\*e}

\begin{document}

\title{Multi-Task Transformer with uncertainty modelling for Face Based Affective Computing}

\author{Gauthier Tallec\\
ISIR\\
\and
Jules Bonnard\\
ISIR\\
\and
Arnaud Dapogny\\
Datakalab\\
\and
Kevin Bailly\\
ISIR/Datakalab\\
\\
}

\maketitle

\begin{abstract}

Face based affective computing consists in detecting emotions from face images. It is useful to unlock better automatic comprehension of human behaviours and could pave the way toward improved human-machines interactions. However it comes with the challenging task of designing a computational representation of emotions. So far, emotions have been represented either continuously in the 2D Valence/Arousal (VA) space or in a discrete manner with Ekman's 7 basic emotions (FER). Alternatively, Ekman's Facial Action Unit (AU) system have also been used to caracterize emotions using a codebook of unitary muscular activations. ABAW3 and ABAW4 Multi-Task Challenges are the first work to provide a large scale database annotated with those three types of labels.
In this paper we present a transformer based multi-task method for jointly learning to predict valence arousal, action units and basic emotions. From an architectural standpoint our method uses a taskwise token approach to efficiently model the similarities between the tasks. From a learning point of view we use an uncertainty weighted loss for modelling the difference of stochasticity between the three tasks annotations.

\end{abstract}

\section{Introduction}
The Affective Behavior Analysis In-The-Wild (ABAW) competitions are a serie  of machine learning based affective computing challenges based on the Aff-Wild2 database \cite{zafeiriou2017aff, kollias2019deep}. In the past few years,  the competition objectives have focused on learning to predict different face expression representation \cite{kollias2020analysing, kollias2021analysing} either jointly \cite{kollias2019face, kollias2021affect, kollias2021distribution, kollias2017recognition} or in a separate manner \cite{kollias2022abaw3}.  This year is no exception, as ABAW4 proposes (a) A challenge for Multi-Task learning that consists in predicting Valence-Arousal, Facial Expression, and Action Units \cite{kollias2022va} (b) A challenge for learning to predict the 7 basic emotions from synthetic data \cite{kollias2022learn, kollias2018photorealistic, kollias2020va, kollias2020deep}.

We participate in the multi-task learning challenge because this is one of the first work to provide a large scale database (170m frames) annotated in all three emotion representations (AU, VA and FER).

One of the main challenge when learning to predict emotion representations is that the provided data are often noisy due to the intrinsic subjectivity of the annotation task. It is all the more a problem when trying to jointly predict several emotion representations as those representations may display different level of noise due to their different level of subjectivity. For example Valence Arousal annotations are slightly noisier than Action Units due to the continuous nature of the annotations. To tackle that challenge we leverage a classic multi-task transformer architecture that we supervise with an uncertainty weighted loss \cite{kendall2018multi} that takes into account the different stochasticity levels of each task.

\section{Related Work}

The lack of annotated datasets in different emotional representations make multi-task face analysis models rarer in the literature. Nonetheless, the recent ABAW challenges have increased the interest in the multi-task approach to predict the three emotion descriptors.

For the ABAW3 challenge, Didan Deng \cite{winnerabaw} introduced a "Sign-and-Message" model which uses an Emotion Transformer to predict Action Units and metric learning to predict facial expressions, valence and arousal.

Jeong \textit{et al.}, the runner-ups of the ABAW3 Multi-Task challenge jointly leveraged the video data with an LSTM and the audio data with a SoundNet. These latent representations are concatenated and fed to different task classifiers and regressors. A task discriminator is used to improve the individual task predictions.

However, the ABAW4 challenge doesn't provide any audio data and the available video data and the important number of missing frames in each video may prevent the use from of temporal methods.
\section{Methodology}

In this section we build up our transformer architecture based on the background material provided in \cite{tallec2022multi}. Furthermore we provide ourselves with a multi-task dataset $\mathcal{D} = \{ \*x_{i}, \va_i, \au_i, \expr_i\}_{i=1}^{N}$ where $\*x\in\Real^{H\times W \times C}$ is the input image, $\va\in\Real^{2}$ is the valence arousal ground truth, $\au\in\{0, 1\}^{12}$ is the AU ground truth and $\expr\in\{0, 1\}^{8}$ is the basic emotion where the $8^{\text{th}}$ label features the additional 'Other' class that is present in ABAW4.

\subsection{General Architecture}

ViT-like architecture are known to be sensitive to overfitting and to therefore require a lot of reliable data to be able to overcome the performance of basic convolutional networks.

For that purpose we adopt an hybrid architecture in which we first feed the input image to a convolutional encoder which consist in a convolutional encoder in which the final pooling layer is replaced by a 1D convolution with output dimension $d$.

Each of the $H' \times W'$ output patch representations are then added an usual cos-sin based encoding to account for its position in the image. Those patches are then passed through $N_x$ classic self attention layers resulting in a patch based image represention that we denote $\mathbf{p} \in \Real^{H' \times W' \times d}$ the resulting image representation.

To predict each of the three tasks we provide ourselves with three learned task tokens. Those tokens are then fed to a sequential arrangement of $N_t$ attention modules. Each cross attention module is composed of two main block:
\begin{enumerate}
    \item The first block consist in self-attention between tokens so that each task can learn to exploit the information from the other tasks.
    \item The second block is a cross-attention step in which each token learns which part of the $\*p$ is most useful to predict its associated task.
\end{enumerate} 

At the end of the $N_t$ modules, each refined token is fed to a sequence of $N_d$ task specific dense layers where all activations are relu except for the last one which is linear. 

The final output of this process is a prediction $\hat{\*v}$ for valence arousal, and logits $\*u_{\*a}$ and $\*u_{\*e}$ for action units and emotion respectively. The predictions for actions units $\hat{\*a}$ and emotion $\*e$ are computed using different temperature $T_{\*a}$, $T_{\*e}$ to account for the different level of stochasticity of those two tasks:
\begin{equation}
    \hat{\*a} = \sigma(\frac{\*u_{\au}}{T_{\au}}), \hat{\expr} = S(\frac{\*u_{\expr}}{T_{\expr}}),
\end{equation}
where $\sigma$ and $S$ denotes sigmoid and softmax function respectively.

\subsection{Uncertainty Based Supervision}

To account for task stochasticy we assigned temperatures $T_{\*a}$ and $T_{\*e}$ to the action unit and emotion prediction task respectively. Similarly we assign a variance matrix to arousal and valence prediction. To simplify we work under the assumption that this variance matrix is diagonal and that valence and arousal have the same intrinsec variance $\sigma^{2}_{\va}$. The final Maximum likehood based supervision loss is then :
\begin{equation}
    \mathcal{L} = \mathcal{L}_{\va} + \mathcal{L}_{\au} + \mathcal{L}_{\expr}
\end{equation}
where : 
\begin{align}
    \mathcal{L}_{\va} &= \frac{1}{2\sigma^{2}_{\va}} \text{MSE}(\hat{\va}, \va),\\
    \mathcal{L}_{\au} &= \frac{1}{2T_{\au}} \text{BCE}(\*u_{\au}, \au), \\
    \mathcal{L}_{\expr} &= \frac{1}{T_{\expr}} \text{CCE}(\*u_{\expr}, \expr),
\end{align}
and MSE, BCE and CCE denotes mean squared error, binary crossentropy and categorical crossentropy respectively. In practice, as all three labels are not present for each examples we mask the losses for the missing labels and reweight the total loss with the number of present labels so that each example get the same contribution in the loss.

\label{sec:methodology}
\section{Experiments}

\subsection{Implementation Details}
For all our experiments we use a Resnet50 pretrained on VGGFACE2 as the convolutional backbone. For each resized image of size $224\times224\times3$, it outputs a $7\times7\time 2048$ patch based representation that we project in $d=768$ dimensions using a 1D convolution. We then apply $N_x = 2$ self attention layers with $12$ attention heads to get the final image representation $\mathbf{p}$.

As far as the task-token based part is concerned it consists in $N_t = 2$ cross attention modules with $12$ attention heads followed by $N_d = 4$ taskwise dense.

The variance $\sigma^{2}_{\va}$ of valence arousal is set to $1$. The temperature of action units and emotion are respectively set to $T_{\au} = 1$ and $T_{\expr} = 5$. 

\subsection{Data Augmentation}
ABAW4 dataset features in the wild data, so that there is a lot of variability in images and potentially between the train, the valid and the test set. To reduce the impact of those variabilitities we use classical geometrical data augmentation : random rotations, translations and zooms. Furthermore we tried using multi-task MixUp which consist in feeding random convex combinations of input images and supervising each task with the same random convex combination of each task ground truth. 

\subsection{Temporal Smoothing}
The provided test set consists in images sampled from videos. To handle the dynamic of labels we simply compute the final predictions as temporal means of predictions in windows of size $S=30$. For valence/arousal, the predictions are directly smoothed. For AU and Emotions, the logits are smoothed and the appropriate activation function is applied on the smoothing output. Missing images in the videos are simply ignored in the mean computation.

Table \ref{tab:abaw4_submissions} summarize the validation scores of our two submissions. Our third submission is simply the mean of the two first submission.

\begin{table}
\centering
    \label{tab:abaw4_submissions}
    \resizebox{\columnwidth}{!}{\begin{tabular}{|c|c|c|c|c|c|c|c|}
\hline
ID & Mixup & AU F1 & Emotion F1 & VA CCC & ABAW4 \\ 
\hline
1 & $\times$ & 0.25 &	0.48 &	0.39 & 	1.12 \\
\hline
2 & \tick & 0.27 & 0.47 & 0.38 & 	1.11 \\
\hline
\end{tabular}}
    \caption{Summary of validation scores for submissions to the ABAW4 Challenge}
\end{table}

\section{Conclusion}
In this work we made use of a multi-task transformer for jointly predicting arousal/valence, basic emotions and action units. We further refined our method by modelling the different levels of stochasticity for each task. Our method finished at the $8^{\text{th}}$ position of the ABAW4 Challenge.

\section{Acknowledgements}
This work was granted access to the HPC resources of IDRIS under the allocation 2021-AD011013183 made by GENCI.

{\small
\bibliographystyle{ieee_fullname}
\bibliography{main}
}

\end{document}